# SPONTANEOUS EXPRESSION CLASSIFICATION IN THE ENCRYPTED DOMAIN

*Segun Aina[1], Yogachandran Rahulamathavan[2], Raphael C.-W. Phan[1], Jonathon A. Chambers[1]*

[1]Advanced Signal Processing Group, School of Electronic, Electrical and Systems Engineering, Loughborough University, UK
{S.Aina, R.Phan, J.A.Chambers} @lboro.ac.uk

[2]School of Engineering and Mathematical Sciences, City University London, UK
yogachandran.rahulamathavan.1@city.ac.uk

*Abstract–* To date, most facial expression analysis have been based on posed image databases and is carried out without being able to protect the identity of the subjects whose expressions are being recognised. In this paper, we propose and implement a system for classifying facial expressions of images in the encrypted domain based on a Paillier cryptosystem implementation of Fisher Linear Discriminant Analysis and k-nearest neighbour (FLDA + kNN). We present results of experiments carried out on a recently developed natural visible and infrared facial expression (NVIE) database of spontaneous images. To the best of our knowledge, this is the first system that will allow the recognition of encrypted spontaneous facial expressions by a remote server on behalf of a client.

## 1. INTRODUCTION

Facial expressions are the changes in the face stimulated by a person's emotional state and are one of the intuitive ways in which humans communicate their emotions. The classification of facial expressions allows the identification of such emotions, and forms an integral part of affective computing which is computing that relates to, arises from, or deliberately influences emotion or other affective phenomena. As such, there is an on-going research interest in the automated recognition of these expressions by computer systems within the areas of pattern recognition, human-computer interaction, human cognition and behavioural science [1].

Most of the existing research in the area of expression recognition and emotion inference is based on posed expression databases which are stimulated by requesting that subjects perform a sequence of emotional expressions in front of a camera. These artificial expressions are usually exaggerated. On the other hand, spontaneous expressions may be subtle and vary in intensity from subject to subject. They will also often differ from posed expressions in both manner and timing. As such, further to the need to infer emotion is the need to do so on a natural database thus moving from artificial to natural expression recognition [1], effectively leading to more practical applications thereof.

Moreover, there is increasing need to outsource computational processes while maintaining privacy, which has very recently prompted research in the area of facial expression classification in the encrypted domain. For example, an advertiser may wish to identify the expressions of consumers in order to estimate their affective responses and reactions to advertising campaigns. This could be done using an expression database on a remote server hosted by a (potentially untrustworthy) third-party provider, in which case the identities of the consumers will need to be kept private (as it is impractical to seek the approval of every consumer who views an advertisement). In this example, the third-party provider who hosts the expression database and the advertiser who wishes to recognise the expression of their consumers can be referred to as the server and client respectively.

To achieve this, the homomorphic properties of a public-key based Paillier cryptosystem will be leveraged in order to keep the images of the subjects encrypted throughout the exchange between the server and the client [2-4] while obtaining the same levels of accuracy as can be obtained on plain (non-encrypted) images.

The rest of the paper will be arranged as follows: Section 2 and 3 will discuss the process of facial expression recognition using Fisher linear discriminant analysis (FLDA) and k-nearest neighbour (k-NN) in the plain domain (PD) and in the encrypted domain (ED) respectively. In Section 4, we will describe the experimental setup and analyse the performance of the algorithm in Section 5. The inferred conclusions are presented in Section 6.

## 2. FISHER LINEAR DISCRIMINANT ANALYSIS FOR EXPRESSION RECOGNITION

The process of automatic expression classification as implemented in this paper involves two steps the first of which is feature extraction using Fisher linear discriminant analysis (FLDA) [5]. FLDA is a well-known dimensionality reduction tool based on principal component analysis (PCA), which extracts a set of key features in order to project a higher dimensional image onto a lower dimensional space. The second step is recognition using a k-nearest neighbour (k-NN) approach which is a basic but effective Euclidean distance classifier [6] that matches the expression of a projected test image against a set of projected training images such that the test image is allocated to the same expression class as the training image to which it is closest. Section 2.1.1 describes this more formally.

## 2.1. FLDA in the Plain Domain

The matrix representation of a grayscale image (in which each element in the matrix represents a corresponding pixel value within the image) can be concatenated into a one-dimensional vector.

Given $M$ training images to be used to determine the lower dimensional feature space when concatenated into vectors of dimension – $n$ is given as $\{x_1, x_2, ..., x_M\}$, that is $x_i \in \mathcal{R}^{n \times 1}\ \forall_i$. First, the vectorized training images need to be mean centered and this can be achieved by subtracting the vector representing the mean of all the training images $\bar{x}$ from each image vector $x_i$, where $\bar{x}$ can be evaluated as: $\bar{x} = \frac{1}{M}\sum_{i=1}^{M} x_i$. The new (lower) dimensional feature space vector $y_i$ corresponding to image vector $x_i$ is obtained by the following linear projection:

$$y_i = W_{opt}^T (x_i - \bar{x}), \quad i = 1, ..., M \quad (1)$$

where $(x_i - \bar{x})$ are the mean centered images, and $W_{opt}$ is an optimum projection matrix with orthonormal columns (with $(\cdot)^T$ denoting the transpose). The optimum projection matrix is given by [5]:

$$W_{opt}^T = [W_{flda}^T \cdot W_{pca}^T] = [w_1, w_2 ... w_m] \quad (2)$$

where $[w_1, w_2 ... w_m]$ denote the feature vectors obtained from both $W_{flda}$ and $W_{pca}$. In order to evaluate $W_{pca}$, consider that the principal component (PC) vectors are the eigen vectors of the covariance matrix $S_T$ (scatter matrix), where $S_T$ is defined as:

$$S_T = \sum_{i=1}^{M} (x_i - \bar{x})(x_i - \bar{x})^T \quad (3)$$

Using eq. (1) and (2), the total covariance matrix of feature vectors $\{y_1, y_2, ..., y_M\}$ is $W^T S_T W$. The optimal PCA projection matrix $W_{pca}$ is selected to maximise the determinant of the total covariance matrix of projected feature vectors, defined as:

$$W_{pca} = \underset{W}{\mathrm{argmax}} |W^T S_T W| = [w_1, w_2 ... w_m] \quad (4)$$

where $\{w_i \in \mathcal{R}^{n \times 1}\ |\ i = 1, ... m\}$ is the set of eigenvectors relating to the $m$ largest eigenvalues of $S_T$.

For (2), $W_{flda}$ is obtained by maximising the between-class scatter while minimizing the within-class scatter which is calculated as a function of the matrices $S_B$ and $S_W$ respectively. Given that:

$$S_B = \sum_{i=1}^{c} N_i (\mu_i - \bar{x})(\mu_i - \bar{x})^T \in \mathcal{R}^{n \times n} \quad (5)$$

and

$$S_W = \sum_{i=1}^{c} \sum_{x_k \in X_i} (x_k - \mu_i)(x_k - \mu_i)^T \in \mathcal{R}^{n \times n} \quad (6)$$

where $c$ is the number of different classes e.g. $\{X_1, X_2, ..., X_c\}$ representing each expression, $\mu_i$ is the mean of class $X_i$ and $N_i$ is the number of images in that class. As such, $W_{flda}$ can be defined as:

$$W_{flda} = \underset{W}{\mathrm{argmax}} \frac{|W^T W_{pca}^T S_B W_{pca} W|}{|W^T W_{pca}^T S_W W_{pca} W|} \quad (7)$$

### 2.1.1. Classification in the Plain Domain

For recognition, given a vectorized test image $\Gamma \in \mathcal{R}^{n \times 1}$, the test image needs to be mean centered using the mean of the training images $\bar{x}$, subsequently, it is projected onto feature space by:

$$\Omega = W_{opt}^T (\Gamma - \bar{x}) \quad (8)$$

where $(\Gamma - \bar{x})$ is the mean centered test image and $\Omega$ is the corresponding low(er) dimensional feature vector $\Omega = [\Omega_1 ... \Omega_m]^T \in \mathcal{R}^{m \times 1}$ where each element can further be defined as:

$$\Omega_i = W_i^T (\Gamma - \bar{x}), i = 1 ... m. \quad (9)$$

As such, the Euclidean distance $D_i$, between $\Omega$ and $y_i$ for $i=1...M$ can be calculated as:

$$D_i = \|\Omega - y_i\|_2^2, i=1...M \quad (10)$$

the test image projection $\Omega$ is said to belong to the same expression class as the projection of training image $y_i$ for the lowest value of $D_i$.

## 3. ENCRYPTED EXPRESSION RECOGNITION

This section of the paper justifies how the classification of facial expressions can be applied to encrypted images. This is achieved using the principles employed in [2] in which encrypted images of people's faces were recognised by leveraging the homomorphic properties of the Paillier cryptosystem [7]. Using the same principles in addition to a cryptographic protocol for the comparison of two encrypted values, we classify the facial expressions of encrypted images in such a way that it can be done by a server hosted database without revealing the contents of the image to the server.

### 3.1. Paillier Encryption

The Paillier cryptosystem is an additively homomorphic public-key encryption scheme, where its security is based on the decisional composite residuosity problem [7]. For example, given encryption $[\![a]\!]$ and $[\![b]\!]$, for all operations performed with plaintext or cyphertext, the following corresponding encryption can be obtained where $[\![a + b]\!] = [\![a]\!] \cdot [\![b]\!]$. Similarly, multiplying an encryption $[\![a]\!]$ with a constant $c$ can be calculated as $[\![a \cdot c]\!] = [\![a]\!]^c$.

### 3.2. Projection in the Encrypted Domain

A string $Exp_i$ corresponding to the expression class is assigned to the lower dimensional feature vectors $y_i$ for $\{y_1, y_2, ..., y_M\}$ where the server in order to setup the facial expression database obtains $W_{opt} = [w_1, w_2 ... w_m]$ using (2). As a requirement for Paillier encryption, all the individ-ual elements in $W_{opt}$ and $y_i$ ranging 1…M are denoted by integers. To achieve this, elements in $W_{opt}$ are scaled by a factor $S$ and subsequently quantized to the nearest integer. Elements of $y_i$ are simply quantized to the nearest integer. The client (advertiser from earlier example) generates a set of private and public keys, the latter of which is sent to the server (third-party service provider).

$[\![\Gamma]\!]$ is obtained and can be sent to the server when the client encrypts each pixel value of a consumer's facial image $\Gamma$ using the earlier generated public key. At this point, the encryption $[\![\Gamma]\!]$ is obtained using the client's public key; as such neither the server nor anyone else is able to decrypt the image hence keeping the identity of the subject (consumer) completely private and confidential from everyone. The server is able to perform linear operations to determine the expression class e.g. operations (9) and (10) on the encrypted image by leveraging the homomorphic properties of the Paillier cryptosystem described above. The resultant expression class, encrypted by the server using the client's public-key is then sent to the client and is decrypted using their private-key.

Formally, facial expression classification in the encrypted domain requires the evaluation of equations (9), which will be the projection of an encrypted test image and (10), the Euclidean distance measure in order to match the image with an expression class.

For projection of an encrypted test image, equation (9) can be rewritten as:

$$\Omega_i = \sum_{j=1}^{n} w_{i,j}(\Gamma_j - \bar{x}_j) \quad (11)$$

where the following elements from (9) are correspondingly redefined as: $w_i = [w_{1,i}\ w_{2,i}\ ...\ w_{n,i}]^T$, $\Gamma = [\Gamma_1\ \Gamma_2\ ...\ \Gamma_n]^T$ and $\bar{x} = [\bar{x}_i, \bar{x}_i, ..., \bar{x}_n]^T$. On the side of the server, only the encrypted value of a given test image $[\![\Gamma_j]\!]$ is known, as such, homomorphic properties allow the evaluation of encrypted value of $\Omega_i$ (obtained using clients public key), given by:

$$[\![\Omega_i]\!] = \prod_{j=1}^{n}([\![\Gamma_j]\!]\ [\![-\bar{x}_j]\!])^{w_{i,j}} \quad (12)$$

From (10), we obtain the $n$ encrypted values of $[\![\Omega_i]\!]$ that make up the projection of an encrypted test image $[\![\Gamma]\!]$, in order to associate a given test image with an expression class, we compute the encrypted distances $[\![D_i]\!], i = 1,…,M$ between the feature vectors of the test image and the feature vectors of the training images. For this, (10) can be rewritten as:

$$D_i = \sum_{j=1}^{n}(\Omega_j - y_{i,j})^2, i = 1, ..., M,$$

$$= \sum_{j=1}^{n} y_{i,j}^2 + \sum_{j=1}^{n}(-2y_{i,j})\Omega_j + \sum_{j=1}^{n} \Omega_j^2, i = 1, ..., M.$$

Homomorphic properties allow the encrypted distances to be computed as:

$$[\![D_i]\!] = [\![\sum_{j=1}^{n} y_{i,j}^2]\!]\ [\![\sum_{j=1}^{n}(-2y_{i,j})\Omega_j]\!]\ [\![\sum_{j=1}^{n} \Omega_j^2]\!], \quad (13)$$
$$i = 1, ..., M.$$

where the server can obtain $[\![\sum_{j=1}^{n} y_{i,j}^2]\!]$ by encrypting the value $\sum_{j=1}^{n} y_{i,j}^2$ and $[\![\sum_{j=1}^{n}(-2y_{i,j})\Omega_j]\!] = \prod_{j=1}^{n}[\![\Omega_j]\!]^{(-2y_{i,j})}$. The server participates in a two-party computation protocol with the client in order to obtain the value of $[\![\sum_{j=1}^{n} \Omega_j^2]\!]$ as only $[\![\Omega_j]\!]$ is known to the server. During this exchange, the server is also keen to keep the contents of the training database $W_{opt}$ and $y_i$ private. As such, the server additively blinds each feature vector component $[\![\Omega_j]\!]$ with a random element $[\![r_j]\!]$, to obtain $[\![\Sigma_j]\!] = [\![\Omega_j + r_j]\!] = [\![\Omega_j]\!][\![r_j]\!]$ which is sent to the client where it is decrypted to calculate $\Sigma_j^2$ and subsequently $\sum_{j=1}^{m} \Sigma_j^2$ within the plain domain. Once this is done, the client encrypts $[\![\sum_{j=1}^{m} \Sigma_j^2]\!]$ and sends it to the server who then uses it to deduce $[\![\sum_{j=1}^{n} \Omega_j^2]\!]$, given as:

$$[\![\sum_{j=1}^{n} \Omega_j^2]\!] = [\![\sum_{j=1}^{m} \Sigma_j^2]\!] \cdot \prod_{j=1}^{m}\left([\![\Omega_j]\!]^{-2r_j}[\![-r_j^2]\!]\right).$$

In doing so, the server has calculated the encrypted distances in (13). The next step in associating a test image with an expression class is to identify the image corresponding to the lowest encrypted distance.

### 3.3. Classification in the Encrypted Domain

The objective is to establish the lower of two encrypted $l$-bit values $[\![D_i]\!]$ and $[\![D_j]\!]$. The server calculates $[\![z_{i,j}]\!] = [\![2^l + D_i - D_j]\!] = [\![2^l]\!][\![D_i]\!][\![D_j]\!]^{-1}$, where $z_{i,j}$ is a positive $(l+1)$-bit value. Let the most significant bit of $z_{i,j}$ be represented as $\tilde{z}_{i,j}$, then $\tilde{z}_{i,j} = 0 \Leftrightarrow D_i < D_j$ and $\tilde{z}_{i,j} = 2^{-l} \cdot (z_{i,j} - (z_{i,j} \bmod 2^l))$. Homomorphic properties allow $\tilde{z}_{i,j}$ to be calculated as $[\![\tilde{z}_{i,j}]\!] = ([\![z_{i,j}]\!][\![z_{i,j} \bmod 2^l]\!])^{-2^{-1}}$. The server needs to engage the client to calculate $[\![z_{i,j} \bmod 2^l]\!]$ as only $[\![z_{i,j}]\!]$ is known. As previously done, the server generates and applies a random blinding value as $[\![z_{i,j} + r]\!] = [\![z_{i,j}]\!][\![r]\!]$ which is sent to the client. Once received, the blinded value is decrypted and $z_{i,j} + r \bmod 2^l$ is reduced. The result is encrypted and sent back to the server who retrieves it as:

$[\![z_{i,j} \bmod 2^l]\!] = [\![z_{i,j} + r \bmod 2^l]\!][\![r \bmod 2^l]\!]^{-1}$. Again using a collaborative two-party calculation protocol, the server obtains the encrypted minimum as $[\![\tilde{z}_{i,j} \cdot (D_i - D_j) + D_j]\!]$ and the encrypted expression class matching that minimum distance, given by $[\![\tilde{z}_{i,j} \cdot (Exp_i - Exp_j) + Exp_j]\!]$. This is then returned to the client who decrypts it to find the expression class of the test image.

## 4. EXPERIMENTAL SETUP

Experiments were performed on the spontaneous database of the Natural Visible and Infrared facial Expression NVIE database [1], which was developed by using videos to stimulate expressions for recognition and emotion inference. The developers of the database concede that not all subjects displayed the emotions typically used in expression classification i.e. anger (AN), disgust (DI), fear (FE), happiness (HA), sadness (SA), surprise (SU) and neutral (NE) in some cases. As such, only the three expressions that they deemed to have been successfully elicited were used in experiments, i.e. DI, FE and HA. From the visible database, a subset of 311 expression images was collected for the 3 classes from the apex folder. Using bootstrapping, these were divided into ten subsets of 72 images (24 images/class), without allowing the same subject to appear multiple times in the same subset. The averaged performance results are provided in Section 5. Pre-processing included manual eye alignment/cropping and each image was converted to greyscale and sized as 90x90. A leave-one-out strategy was used for cross validation.

## 5. PERFORMANCE REVIEW

The performance of the algorithm is measured using the experimental setup described in Section 4 and the following results were obtained by averaging the (percentage) results from the 10 subsets.

Table 1: Confusion matrix of average results (%).

| %   | DI    | FE    | HA    |
|-----|-------|-------|-------|
| DI  | **66.67** | 29.17 | 4.17  |
| FE  | 29.17 | **58.33** | 12.50 |
| HA  | 8.33  | 8.33  | **83.33** |
| Av. |       | **69.44** |   |

These results obtained using PCA+LDA (FLDA) are shown to be better than results obtained using other feature extraction methods namely PCA, Active Appearance Model (AAM), and AAM+LDA [5].

Table 2: Results of other feature extraction methods (%)[1].

| %   | PCA   |       |       | AAM   |       |       | AAM+LDA |       |       |
|-----|-------|-------|-------|-------|-------|-------|---------|-------|-------|
|     | DI    | FE    | HA    | DI    | FE    | HA    | DI      | FE    | HA    |
| DI  | 50.60 | 31.33 | 18.07 | 65.06 | 27.71 | 7.23  | 59.04   | 30.12 | 10.84 |
| FE  | 27.42 | 50.00 | 22.58 | 30.65 | 53.23 | 16.13 | 37.10   | 45.16 | 17.74 |
| HA  | 14.29 | 14.28 | 71.43 | 13.19 | 6.59  | 80.22 | 12.09   | 12.09 | 75.82 |
| Av. | 58.47 |       |       | 67.80 |       |       | 61.44   |       |       |

The classification results (%) obtained in the encrypted domain ED increases, as value of the scaling factor $S$ is increased up to the maximum percentage obtained in the plain domain PD as shown in Table 3 below.

Figure 1: Examples of pre-processed facial expression images (top row) and encrypted equivalent (bottom row).

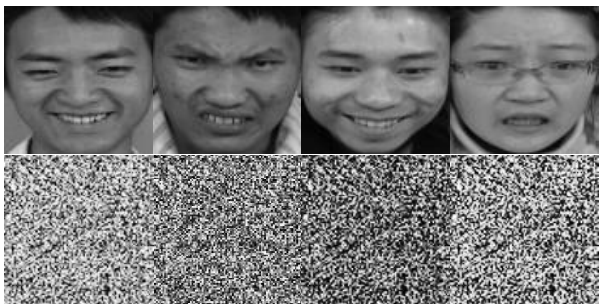

Table 3: Scaling factor and corresponding classification accuracies (%) of a sample dataset.

| ED Scaling factor | $S = 1$ | $S = 10^1$ | $S = 10^2$ | $S = 10^3$ | $S = 10^4$ |
|---|---|---|---|---|---|
| Accuracy (%) | 16.67 | 63.89 | 68.06 | 68.06 | 68.06 |
| PD (%) | **68.06** | | | | |

## 6. CONCLUSION

We propose and implement an encrypted domain-based automatic spontaneous expression recognition system using FLDA. It leverages the homomorphic properties of Paillier encryption to allow classification of facial images while protecting the identities of the subjects in the images at all stages of the classification process. The algorithm is evaluated using a spontaneous dataset and shows classification can be carried out in the encrypted domain without compromising the classification accuracies obtained in the plain domain.